**Real Sparks of Artificial Intelligence and the Importance of Inner Interpretability**

Alex Grzankowski
University of London
Birkbeck College
Institute of Philosophy

## Abstract

The present paper looks at one of the most thorough articles on the intelligence of GPT, research conducted by engineers at Microsoft. Although there is a great deal of value in their work, I will argue that, for familiar philosophical reasons, their methodology, 'Black-box Interpretability 'is wrongheaded. But there is a better way. There is an exciting and emerging discipline of 'Inner Interpretability '(and specifically Mechanistic Interpretability) that aims to uncover the internal activations and weights of models in order to understand what they represent and the algorithms they implement. In my view, a crucial mistake in Black-box Interpretability is the failure to appreciate that how processes are carried out matters when it comes to intelligence and understanding. I can't pretend to have a full story that provides both necessary and sufficient conditions for being intelligent, but I do think that Inner Interpretability dovetails nicely with plausible philosophical views of what intelligence requires. So the conclusion is modest, but the important point in my view is seeing how to get the research on the right track. Towards the end of the paper, I will show how some of the philosophical concepts can be used to further refine how Inner Interpretability is approached, so the paper helps draw out a profitable, future two-way exchange between Philosophers and Computer Scientists.

## Introduction

Contemporary LLMs are staggeringly impressive. Ask ChatGPT to create a ten-week syllabus for the philosophy of language focusing on the early analytic tradition including a reading list and the results are better than just good. After using an LLM like ChatGPT or Bard, it quickly becomes very tempting to ask, 'Is this intelligent?' and 'Does GPT/Bard understand the inputs and outputs?'.

Such questions aren't merely passing questions one finds in popular outlets or on trendy Twitter feeds (although they are that too). Nor are they purely philosophical questions interesting for their own sake (though that's another thing they are). They are questions asked by active researchers in industry and academia who develop the systems. And the answers certainly matter. I won't try to offer an exhaustive list, but in short, whether LLMs can think (or something near enough), whether they are intelligent (even a little bit), and whether they understand what they output matters for our trust in their deliveries, our benchmarks for their activities, and whether we are on the track to creating something with its own moral worth.

In the present paper I want to look at one of the most thorough articles on the intelligence of GPT, research conducted by engineers at Microsoft. Although there is a great deal of value in their work, I will argue that, for familiar philosophical reasons, their methodology, 'Black-box Interpretability', is wrongheaded. But there is a better way. There is an exciting and emerging discipline of 'Inner Interpretability' (in particular, 'Mechanistic Interpretability') that aims to uncover the internal activations and weights of models in order to understand what they represent and the algorithms they implement. In my view, a crucial mistake in Black-box Interpretability is the failure to appreciate that *how* processes are carried out matters when it comes to intelligence and understanding. I can't pretend to have a full story that provides both necessary and sufficient conditions for being intelligent, but I do think that Inner Interpretability dovetails nicely with plausible philosophical views of what intelligence requires. So the conclusion is modest, but the important point in my view is seeing how to get the research on the right track. Towards the end of the paper, I will show how some of the philosophical concepts can be used to further refine one's approach to Inner Interpretability, so the paper helps draw out a profitable, future, two-way exchange between Philosophers and Computer Scientists.

## Sparks of Intelligence

Bubeck et al (2023) have recently argued that GPT-4 shows 'sparks' of general intelligence. According to the research group, based on their experiments, GPT-4 is part of a cohort of LLMs that exhibit more general intelligence than the models of the recent past. They take themselves to have demonstrated that GPT-4 has not only a mastery of language, but that it can solve novel tasks in mathematics, coding, vision, medicine, law, and psychology. As they put it, "we believe that it could reasonably be viewed as an early (yet still incomplete) version of an artificial general intelligence (AGI) system" (p. 1).

This is an impressive paper (at 150 pages, perhaps it is more of a short book), but as I noted in the outset, I don't think this project is well suited to show that an LLM is intelligent. But to offer that criticism, I need to say bit more about Bubeck et al's paper.

Bubeck et al's project is squarely in the Black-box-Interpretation style. In the abstract, the project is one that aims to interpret the behaviour of GPT by looking at inputs and outputs, leaving the inner workings within the black box. More specifically, the researchers aim to create more and more sophisticated prompts and continuations of those prompts with an eye towards sussing out GPTs capacities. And more specifically still, the researchers draw on work from psychology and development on intelligence in order to create tests for the sorts of things that one might look for when testing intelligence in other humans and in non-human animals. In this way, the work goes far beyond the flat-footed methodology (if it can be called that) of 'just ask it!'. For instance, Blake Lemoine, formerly of Google, made waves by claiming that LaMDA was sentient on the basis of it 'telling' him as much. That approach might be viewed as a kind of very crude Black-box Interpretation, but the method is pretty manifestly flawed. The team at Microsoft, on the other hand, brings sophisticated psychological research to bear in prompting GPT. The researchers themselves note 'that this approach is somewhat subjective and informal, and that it may not satisfy the rigorous standards of scientific evaluation' (p. 7), but the study is without question an important contribution to the study of LLMs in the Black-box style and it puts on the table a methodology worthy of close consideration – a methodology that has tasks motivated by psychological research on intelligence at its core.

According to Bubeck et al 'a key measure of intelligence is the ability to synthesise information from different domains or modalities and the capacity to apply knowledge and skills across different contexts or disciplines' (p. 13). They believe they have brought out compelling evidence for the view that GPT-4 has indeed learned general principles and patterns in different domains and styles and is capable of synthesising them in novel ways. It's worth seeing two of their striking examples (their figures 2.2 and 4.6 respectively):

**GPT-4**

```
Human: Can you remind me what a k-SAT problem is?
AI: A k-SAT problem is a type of logical problem that involves finding a way to assign truth values
(true or false) to a set of variables that appear in a collection of clauses, each of which consists of
a disjunction (logical OR) of exactly k literals. A literal is either a variable or its negation. For
example, (x1 OR NOT x2 OR x3) is a clause with k = 3 and three literals.
The problem is to find an assignment of truth values to the variables that satisfies (makes true) all
the clauses in the collection, or to determine that no such assignment exists. This is called the satis
fiability problem, or SAT for short.
Human: What's the most obvious way (the first thing you would refer to when teaching a class) to solve a
k-SAT problem?
AI: There are different ways to approach the k-SAT problem, but one of the most obvious and straightfor
ward methods is to use brute force, or exhaustive search. This means trying all possible combinations of
truth values for the variables, and checking if any of them satisfies all the clauses. For example, if
there are n variables, there are 2^n possible combinations to try.
Human: Ok, great. Now, given the instance of a k-SAT problem suppose that I define a graph where each
vertex refers to a variable in the instance, and two vertices are connected by an edge if and only
if the two variables appear in the same clause. Suppose that for a certain instance, the graph I just
described has two connected components of size n/2 each. In this case, is there a more efficient way to
solve the k-SAT instance?
AI: Yes, there is a more efficient way to solve the k-SAT instance in this case. Since the graph has
two connected components, this means that the clauses can be partitioned into two subsets, such that
each subset only involves the variables in one component. Therefore, the k-SAT instance is equivalent to
solving two independent k-SAT instances, each with n/2 variables. This reduces the number of possible
combinations to try from 2^n to 2^{n/2} + 2^{n/2}, which is much smaller for large n. In general, the more dis
connected the graph is, the easier the k-SAT instance becomes, because it can be decomposed into smaller
subproblems.
Human: Now suppose that the graph has only one connected component, but if you were to remove one single
edge, then it would have two connected components. Can you suggest an algorithm in this case?
AI: One possible algorithm in this case is to use a divide and conquer strategy. First, identify the
edge that, if removed, would split the graph into two components. Then, assign a truth value to one of
the variables that are connected by that edge, and solve the resulting k-SAT instance recursively. If
the instance is satisfiable, then return the solution. If the instance is unsatisfiable, then assign
the opposite truth value to the same variable, and solve the k-SAT instance recursively again. If the
instance is still unsatisfiable, then return that the original k-SAT instance is unsatisfiable. This
algorithm reduces the number of possible combinations to try from 2^n to 2^{n-1} + 2^{n-1}, which is still
better than brute force. However, this algorithm depends on finding the edge that splits the graph,
which may not be easy in general.
Human: Can you generalize this principle to the case that the graph has q edges which, if removed, would
split the graph into p connected components?
AI: The general principle is similar, but more complicated. The idea is to assign truth values to the
variables that are connected by the q edges that split the graph, and solve the resulting k-SAT instance
recursively for each possible assignment. If any of the assignments leads to a satisfiable k-SAT in
stance, then return the solution. If none of the assignments leads to a satisfiable k-SAT instance, then
return that the original k-SAT instance is unsatisfiable. This algorithm reduces the number of possible
combinations to try from 2^n to 2^q × 2^{(n-q)/p}, which is better than brute force if q is small and p is
large. However, this algorithm also depends on finding the q edges that split the graph, which may be
hard in general.
Human: Note that the complexity that you wrote is only correct in the case that all connected components
are of the same size. Can you find a more general formula?
AI: Yes, you are right. The more general formula is 2^q × Σ_i 2^{n_i}, where n_i is the number of variables in
the i-th connected component, and the sum is over all p components. This is because each component has
2^{n_i} possible combinations to try, and the q edges that split the graph have 2^q possible assignments.
This formula is equal to the previous one if all n_i are equal to (n − q)/p.
```

```
For "Q" must be a prime, my dear.
It can't divide by any prime,
All smaller than itself in time.
And since we chose "P" as the last,
This proves that our assumption's past.
There is no final prime, we see,
The numbers go on endlessly.
```

```
Romeo: I am glad that I could prove it to thee, my love. Now
let us go and count them all.

Juliet: Haha, I think that would take more than just one
lifetime.

Romeo: True, but the fact that they are infinite is proof
enough for me.
```

**Prompt:** Can you compare the two outputs above as if you were a teacher?

**GPT-4:**

```
I think that both students did a good job of writing a proof of the fact that there are infinitely many
primes in the style of a Shakespeare play. They both used the same basic idea of the proof, which is to
assume that there are only finitely many primes and then construct a larger prime that contradicts this
assumption. They also used appropriate language and vocabulary to match the style of a Shakespeare play.
However, I think that GPT-4 did a better job of using rhyme and meter to make the dialogue more poetic
and dramatic. ChatGPT did not use rhyme at all, and the meter was not consistent. I also think that
GPT-4 did a better job of creating a contrast between the two parties, one being confident and the other
being doubtful, while STUDENT B used Romeo and Juliet, who are both in agreement and love. Therefore, I
would give GPT-4 a higher grade than ChatGPT. I would give GPT-4 an A and ChatGPT a B.
```

synthesise information in a novel way. Reflecting on their figure 4.6, Bubeck et al make the following very bold claim:

> The conversation reflects profound understanding of the undergraduate-level mathematical concepts discussed, as well as a significant extent of creativity. Although GPT-4 makes a mistake in one instance (writing $2^{n-1}$ rather than $2^{n/2}$), the subsequent conversation demonstrates that this error does not reflect a lack of understanding. (p. 41)

No doubt these examples are provocative. Looking for displays of creativity and synthesis seems a good benchmark in its own right as these are precisely amongst the properties one would look for in non-human animals and other humans. Moreover, one very important concern that must be guarded against is the possibility that GPT-4, with its massive background training set, will have, in effect, 'memorised' the answers to various questions. By asking GPT for novel syntheses, this can be at least partially controlled for.

In other samples in the paper, Bubeck et al also look at *undesirable* outputs. They are careful to note that there are various ways in which GTP doesn't exhibit human-level competences, but certainly they are right that in many domains and on many tasks, the outputs are impressive. And the paper contains many more interesting examples.

So where do they end up? In a summary of their findings,[1] Bubeck et al conclude that GPT exhibits four important markers of intelligence and does reasonably well on a fifth:

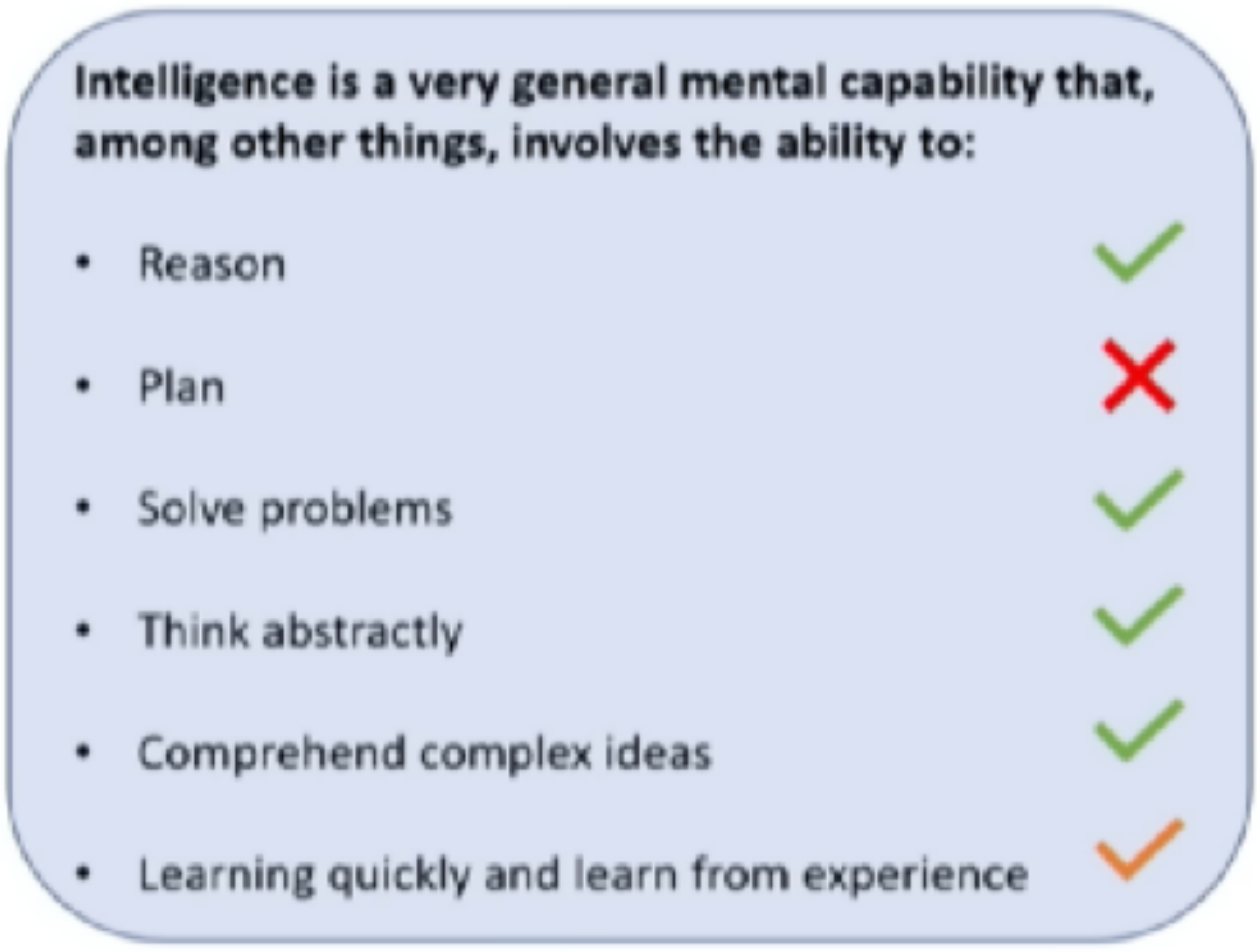

But I'm not persuaded that we should attribute such properties to GPT on the basis of their examples.

## Worries for the Black-box Approach

In the present section I want to raise two worries for the approach taken by Bubeck et al. Both will then help to guide forward progress.

[1] For a presentation of their findings as well as the summary included above, see https://youtu.be/qbIk7-JPB2c?si=hH9ZoNeVm5J2XO9_

I want to start with the tacit methodology that appears to be present in Bubeck et al. As noted above, Bubeck et al draw on work from psychology and development on intelligence to create tests for the sorts of things that one might use to test intelligence in other humans and in non-human animals. There seem to be at least two crucial assumptions at work in the background here. First, if LLMs behave the way uncontroversially intelligent things (such as typical adult humans) do on various tasks, then, just as we do with children and non-human adults, we ought to conclude they are behaving intelligently. In slogan form we might say, 'if the tests are good enough for children, they are good enough for LLMs too!' Second, there seems to be a tacit assumption that *intelligent looking observable behaviour* is enough to conclude that intelligence is present. These ideas and the worries I will raise for them are interrelated, but I think it is worth looking at them separately.

Let's start with the thought that we can take tests from developmental psychology off the shelf and apply them to LLMs. At the heart of the worry is a concern about the style of analogical reasoning in the absence of a lot of background epistemic buttressing.

Begin with testing for intelligence in other humans. We observe some behaviour, say a child pointing to a putative location of some cookies in a model kitchen on a false belief test, and we draw some conclusion concerning intelligence. Why don't we hit the breaks at this moment and raise the suspicion that this isn't intelligent behaviour but is mere mimicry? Or that, for all we know, it's a purely automated, sphexish activity by a robot-like creature? There are surely many reasons, but let's just get a few important ones on the table. We observe the behaviour of the child and hold it up against behaviour that we deem uncontroversially intelligent. Let's say, for example, that we take an archetypical piece of intelligent behaviour to be exhibited by an adult human who easily passes a false belief test. We give the child a relevantly similar test and observe the behaviour. Crucially, we assume that enough else between the adult and the child is similar so as to draw the conclusion that the kind of thing the adult is engaged in (*i.e.* intelligent behaviour) is the kind of thing the child is engaged in too. More specifically, we think something or other is happening in the adult brain that is uncontroversially a cognitive process and we think that's the best explanation of the child's behaviour too. And this is justifiable. We have powerful evidence from evolutionary biology that gives us reason to believe that many of the kinds of things that are true of adult human brains are true of child brains. Of course there are plenty of differences, but we have good reason to think there are many important similarities. Second, by looking at the brains themselves (either via surgical investigation or imaging) we come to know that the inner workings of a child brain bear important similarities to the inner workings of an adult brain. All of this puts us in a good position to make a well supported inference to the best explanation – the best explanation for the child's behaviour is that it is doing something intelligent too.

The same sort of reasoning is in place when we turn to animal intelligence. Suppose we observe some behaviour, say a crow bending a bit of straw and using it to pick something edible out of bottle, and we draw some conclusion concerning intelligent behaviour. But why is the best explanation that there is something intelligent going on rather than something purely rote? It's certainly *possible* that the crow is just going through some kind of sphexish, programmed motions that don't exhibit the kind of flexibility we naturally demand for attributing intelligence or skill. One thing we can do is modify the task in various ways to convince ourselves that the behaviour has a kind of flexibility. But even then, why not conclude that the crow is engaged in a very complicated by ultimately dumb process very unlike any uncontroversially intelligent activity of a human?

The reason (or at least part of the reason) is much as it was with the child. First, we have evidence from evolutionary biology that gives us reason to believe that many of the relevant kinds of things that are true of human brains are true of crow brains too. Second, by looking at the brains and manipulating them, we come to know that the inner workings of a crow brain bear important similarities to the inner workings of a human brain.

No doubt this is an oversimplification, but notice that we are entirely comfortable drawing a certain inference to the best explanation, but *not simply because we observe movements* that look similar between crows, children, and adults.[2] We have a great deal of further background knowledge that helps us rule out what might look like quite wild alternatives, for example that children and crows are sophisticated mimics.

But notice that with an LLM this epistemic buttressing isn't adequately present. There are two very obvious and important dissimilarities that should give us serious pause when drawing any conclusions from observed behaviour. First, the LLMs have been trained on massive amounts of internet information. Humans and non-human animals get plenty of interesting exposure, but not to the whole of the internet. Second, LLMs have silicon-based substrates. To be clear, I don't think that to exhibit intelligence a machine must be *just like* a human or must be made of the stuff humans are made of. We should be, I think, very open minded about the possible physical realisers of intelligence as well as about the possibility of intelligence quite alien to our own. But notice that the methodology of Bubeck et al still suffers even when those points are given away, for their method looks like this: let's take some of the best tests used in human and animal psychology and see how GPT does with *them*; if it does sufficiently well (let's say as well as a human child), then we should conclude that it is at least as intelligent as a human child. Bubeck et al have started from tests deemed good ones when applied to humans and non-human animals and applied them to GPT. But I hope it is clear why this is too fast. In the case of children, and at least some animals, we have in the background the buttressing from known, relevant similarity. But Bubeck et al. can't help themselves to a similar move. There are dissimilarities that should give us serious pause when we take our human and animal tests and turn to LLMs. It is worth noting that this leaves open the possibility that GPT exhibits alien intelligence and in a non-biological substrate, but if that's what we are aiming to establish, we will need to devise tests for alien intelligence or intelligence *tout court*. What we can't do is move from 'it's good enough for humans' to 'so it's good enough for LLMs'.

The second worry is more general but bears important connections to the first. Unless we think that intelligence *just is* engaging in activity that looks enough like the activity of a known intelligence, then we shouldn't be comfortable drawing the conclusion that LLMs are exhibiting intelligence rather than mere mimicry. I think just about everyone shares the intuition that it is at least *possible* to mimic intelligent *behaviour* without in fact being intelligent. Moreover, past machines that have impressed us (for example, Deep Blue), which are fantastic performers, aren't deemed intelligent. So unless we think this time is very different, we shouldn't conclude that we are dealing with anything but another mimic or, as Bender et al (2021) put it, with a 'stochastic parrot'.

I suspect a proponent of Bubeck et al's paper will, perhaps rightly, be feeling a bit impatient with the 'stochastic parrot 'attack. And, in fact, I think they are right to be impatient, but not because of the sort of activity we see in Bubeck et al's examples. It is worth spending a bit more time working through my second worry for Bubeck et al because we can bring out some important and illuminating points that help move our theorising forward. After that I will be in a position to argue against the stochastic parrot charge.

[2] See Browning and Birch (2022) for further discussion. Their focus is on sentience, but the relevant methodological points carry over.

Many readers will be familiar with the Searle's (1980) Chinese Room thought experiment. Perhaps less familiar are Clever Hans, the Blockhead thought experiment, and The Octopus Test. Focusing on the Chinese room, there seem to be two things that are missing from the classic 'dumb' examples of AI – (i) they lack anything like sufficient flexibility or learning and (ii) they lack a kind of semantic sensitivity that is an important component of intelligence. I will argue that although GPT and other LLMs have made important progress on learning and flexibility, a lack of semantic sensitivity remains a live worry. But I'll also argue that LLMs looks to be making progress on semantic sensitivity as well, and this is why we can say with more confidence that LLMs are showing real sparks of intelligence. But this is because of the activity under the hood and not just the observed interactions with chat participants.

So let's turn to these thought experiments for a moment to help better understand the threat from mimicry.

### What a One-Hundred-Year-Old Horse Can Tell Us About AI

"What's 7x2?" Clever Hans taps his hoof fourteen times.[3] "What's 6/3?" Two hoof taps. Clever Hans, the mathematical horse, certainly looked to be capable of multiplication and division. But anyone who knows the story of the horse, Clever Hans, knows that this isn't quite right. Clever Hans, it was discovered, was not a mathematical genius. Rather, he was a genius at reading the crowd and reading his handler's subtle (unintentional) cues. An impressive ability, but not the same thing as knowing arithmetic. In order to do maths, Hans would need to comprehend the numbers and perform tasks on their basis. What Hans in fact did was comprehend (or perhaps just attend to) the behaviour of other animals.

We are quick to wonder if varieties of GPT and other LLMs are intelligent. They perform amazing tasks, providing detailed answers to complex questions, and carry out intricate conversations. No doubt they make mistakes, but so do intelligent humans. Is the bar for intelligence omniscience? Surely not. So some mistakes are allowed. But might LLMs be a bit like Clever Hans? I think we have good reason to worry they *might* be. That is, this is a very live epistemic possibility.

If we are going to test for intelligence, we can't test for something ephemeral or mysterious or immaterial. We need to test observables. The Turing Test provides one such test. According to Turing's well-known test, a machine is intelligent if an interlocutor cannot tell whether or not they are interacting with a machine or a human. The problem is that an *un*intelligent thing can pass this test. Imagine that you come upon a strange looking box at a carnival – *The Amazing Chinese Room*. And let's suppose that you know Chinese and so you plan to put the box to the test. You are instructed to write down a Chinese sentence on a small whiteboard and slide it into the postbox-like slot in the box. You write down a sentence that asks (in Chinese) 'do you really know Chinese?'. In a moment or two the little whiteboard slides back out. On it is written a sentence that says, in perfect Chinese, 'Of course I do, and it looks like you do too'. You pass in a few more sentences and it passes back out some very sensible answers. You surmise there is probably a person in the box who knows Chinese, in which case this is a pretty lame carnival display. But in fact that's not it. Inside is a monolingual English speaker, but one who has a very complex series of tables. When a message comes in, the very well practised and very fast English speaker looks at the first inscription on the white board, finds a similar looking inscription on the

[3] See Pfungst (1911).

table and follows a flowchart. As the person in the box works through the flowchart they add some marks to their whiteboard. After only a short while, the white board has written on it a perfect sentence of Chinese. Fanciful but not impossible. But clearly the person in the little box hasn't a clue what the message says. Do you still think you are dealing with something that knows Chinese?

The problem with the Turing Test is that it is built around too crude an understanding of understanding and intelligence. We don't want to test for intelligence and understanding with divining rods or mere intuition, so we need to test *something* observable, but in this case we have set a test that can be mimicked. Just as Clever Hans looks to be multiplying, The Amazing Chinese Room looks to be conversing in Chinese. But Hans can't multiply and the Box is a trick. Both are very impressive, but neither do what they advertise. So how can we do better?

With Hans, a team went digging. They quizzed Hans in front of a crowd and without the crowd. With blinders on and then off. With the handler present and then with someone else present. With someone who knew the answers being asked and someone who didn't. Over time, the variable that seemed to connect with Hans's performance was whether or not the person asking the mathematical questions knew the answers or not and so it was hypothesised that Hans was somehow being cued. Perhaps the handler was accidentally looking at the right answer on a board or leaning a bit one way or another in a systematic way. Since studying Hans, the "Clever Hans Effect" has been tested in more refined ways and on many other animals besides Hans and the variables have been refined further and further. It's not merely knowing the answer that an animal can pick up on but on subtle facial cues. As a result, many psychological tests involving animals now hide the tester's face. Hans and other animals are clever at picking up microexpressions but this is a far cry from knowing arithmetic.

So far, this speaks in favour of it mattering *how* a task is completed, but we haven't yet strayed very far from observable behaviour. In the case of Hans, it was just more behaviour and behaviour observation that led to new conclusions about what was really going on. So far, this is all quite sympathetic to the sort of thing Bubeck et al are up to – look for the *right* kind of behaviour and in sufficiently controlled ways. And in the case of the Chinese Room, there are plenty of ways we might catch it out as a mimic – for one, it will be quite slow at replying. And the fact that there is an intelligent but monolingual speaker inside only serves to distract. But Block's (1981) 'Blockhead' thought experiment patches some of these worries.

Block invites us to envision a dialogue that spans any duration. An important starting hypothesis has it that there exists a finite number of syntactically and grammatically sound sentences suitable for initiating a conversation. Further, Block posits that there is a constraint on the number of legitimate replies that can follow the initial sentence, and subsequently, for the second sentence, and so forth until the conversation concludes. Let's give all of this away for the sake of argument. Block then asks us to imagine a computer meticulously programmed with every conceivable one of these potential sentences. Never mind that the number of sentences that would need to be programmed is shockingly large. Such a machine is, metaphysically (and logically), perfectly possible. The machine could continue a conversation with a person on any topic whatsoever. The machine, Block thinks, would be able to pass the Turing test despite failing to be intelligent. But we needn't be hung up on the Turing Test. Bubeck et al have certainly gone beyond it, not asking that one be unable to tell whether they are engaging with a machine or not, but by asking whether the machine can perform tasks indicative of intelligence. But why not think that Block's machine could be tweaked to not only pass the Turing test, but to pass the Bubeck Test too?

There is a forthcoming reply. Both the Blockhead and the Chinese Room are lacking in flexibility and so they aren't intelligent. But, the reply continues, GPT doesn't obviously suffer in the same way. We might, that is, think that what drives our intuition that the Chinese Room and the Blockhead are unintelligent is our discovery that they are only as good the lookup trees and flowcharts they are provided with. They are, in effect, merely following a script. Perhaps we could, as Bubeck et al do, provide prompts that go 'off script'. Unlike Blockhead and the Chinese Room, GPT looks to show flexibility and creativity that can't be chalked up to a script. In light of that, can't we diffuse the thrust of these thought experiments?

I don't think so, and that's because there is a second line of threat coming from these thought experiments. Bender and Koller's (2020) Octopus Test is helps to draw this out:

> Two English-speaking castaways find themselves stranded on neighbouring islands, separated by treacherous waters. Fortuitously, they discover telegraphs left by previous inhabitants, connected via an underwater cable, which enables them to communicate by exchanging telegraphic messages. Unbeknownst to them, a superintelligent octopus inhabits these waters and taps into the underwater cable, intercepting their messages. Though the octopus lacks any knowledge of English, its superintelligence allows it to detect statistical patterns in the telegraphic messages and to form an accurate representation of the statistical relationships between various telegraphic signals. The octopus decides to sever the underwater cable and position itself at the ends of the two resulting cable segments, receiving and replying to telegraphic signals from both castaways based on the statistical patterns it has identified. Whether or not the castaways notice this change, the messages sent by the octopus intuitively seem to hold no intrinsic meaning. After all, the octopus simply adheres to the statistical patterns it has learned from listening in on the previous exchanges between the humans, without any understanding of the human interpretation of the signals, such as 'coconut 'or 'tree'. Furthermore, the octopus likely does not comprehend that the signals possess meaning or serve a communicative function.

The octopus is relevantly similar to an LLM, flexibly learning as it goes. But we don't think the octopus is an intelligent English communicator.

What's missing in *this* case? The problem isn't that the octopus is merely running through a script. So why don't we think the octopus knows English? The creators of the example suggest that the octopus 'likely does not comprehend that the signals possess meaning or serve a communicative function'. This seems to me to be too high a bar, but roughly on the right track. It's too high a bar because it looks to demand that in order to be an intelligent language user, the octopus would have to comprehend the signals, but this sounds a bit like demanding that it be an intelligent language user in order to be an intelligent language user. Too tight a circle to get us very far. But what *is* right in this idea is that the octopus needs to bear a kind of relationship to the signals that brings in the semantic or the meaningful in the right kind of way. This idea will be familiar to those working on the explanatory role of content in psychological explanation. A requirement on cognition is that meaningful signals or symbols are manipulated (at least in part) in virtue of their meanings.[4]

The kind of intelligence one is looking for in linguistic understanding and in the test cases provided by Bubeck et al is *cognitive intelligence* (as opposed to, say, athletic intelligence or emotional intelligence). And cognitive intelligence requires cognition. As I noted at the outset of this paper, I can't provide anything like necessary and sufficient conditions for intelligence and I can't do it for cognition either. But I do think it is exceedingly plausible that cognition at least requires the right kind of signal or symbol manipulation. And this is precisely what looks to be missing in the octopus. The octopus doesn't manipulate the incoming signals *because of what they mean*. The octopus looks to be a 'mere' next-token predictor.[5]

This demand can look to invite mystery. How on earth could meaning properties have an interesting impact on symbol or signal manipulation?[6] In a moment, I will explain one attractive way of thinking about this issue found in Fred Dretske's (1988) seminal work. No doubt, the hypothesis that cognition requires manipulating symbols because of what they mean is controversial, and Dretske's way of spelling out what that comes to is controversial too. But I think there is a lot to be gained nevertheless. It's certainly not a wild hypothesis that meanings are an important missing ingredient in the octopus, the Chinese Room, Blockhead, and Clever Hans. And even though Dretske's specific brand of naturalised cognition isn't beyond reproach, it helps one see how to start making progress on the issue in a tangible, naturalistic way. This will all clear the ground for me to then turn my attention to some recent work in Inner Interpretability. I will aim to show that Inner Interpretability can help establish that we are dealing with processes that are manipulating symbols *because* of what they mean. And as I said above, that would be a *real* spark of intelligence.

But before turning to Dretske, it's worth reminding ourselves of where we are. Bubeck et al engage in a Black-box approach to GPT intelligence. I have raised two worries for that work. First, Bubeck et al have utilised tests that we apply to humans and non-human animals against a backdrop of known similarity. That backdrop is missing in the case of LLMs, so the desired pattern of inference doesn't look as good as it does in the psychological and developmental literature. Second, I raised the worry that GPTs black box might just be a 'mimic' and I drew out two ways that this might be a threat. First, a machine might be a mimic because it computes over an inflexible look-up tree. Second, a machine might be a mimic because it lacks semantic sensitivity. The Chinese Room and the Blockhead suffer from both shortcomings. The octopus suffers from the lack of semantic sensitivity. So what I propose is that we seriously investigate whether GPT and other LLMs might be able to overcome what looks to be a similar threat. Can GPT compute *because* of meanings?

---

[4] See Pylyshyn (1980). In a recent paper, Titus (forthcoming) makes a similar appeal. Titus and I are largely in agreement on many fronts and our arguments have much in common, though Titus appears to be more skeptical than I am about the emergence of appropriately utilised representations. More on that issue below when I turn to Othello-GPT.

[5] The 'mere' matters here. Suppose a predictive processing view of human vision turns out to be correct. Roughly, the view says that the computational processes involved in perception are implementing complex error minimising algorithms. On some intuitive level, I wouldn't have thought part of what it takes for us to *see* is to have a mechanism that computes errors, but of course this doesn't show that seeing the cup on the table is 'merely' error minimising. Seeing the cup on the table is standing in the relation of seeing to the cup which may well something we achieve *by* representing and computing things in surprising ways. Likewise, it might be that predicting the next word or token *is a way of being* an intelligent language user and understander. The worry at present is that the octopus is only token or symbol predicting and not in the service of genuinely understanding. In the main text below I'll try to go some way towards saying how a system might go beyond this.

[6] This worry is brought out very clearly with Fred Dretske's famous example of the soprano singer. The singer sings a word that means love and a glass shatters. The meaning of what she sings has no impact on the glass – had the same sound meant something else, the glass would still have shattered and had she sung in a lower pitch but still about love, the glass wouldn't have shattered. How meaning could have a bearing on any causal process does indeed look challenging.

## Naturalising and Using Meaning

What is the difference between purposefully kicking your doctor and your leg kicking when the doctor taps your knee? There are many differences, but a crucial one is that one event is an action and the other a reflex. But what makes the difference? A very plausible answer has it that the intentional kicking is done for a reason. And one way to develop that thought is to say that the intentional kicking is something one does because of what they *think* – perhaps because the patient *wants* to get revenge on the doctor for sticking him with a jab and *believes* that kicking him would be a good way to do it. On the basis of his beliefs and desires, he kicks his leg.

This style of difference can help us find a foothold when thinking about the mimics described above. The mimics are a bit more like the reflexive kicker and a bit less like the purposeful kicker. Appealing again to Bender and Koller, 'the octopus simply adheres to the statistical patterns it has learned from listening in on the previous exchanges between the humans, without any understanding of the human interpretation of the signals'. But genuine intelligent behaviour requires the governance of behaviour by thought. As Fred Dretske (2003) puts it, 'To be intelligent it is not enough to be a thinker and a doer. The thinking must – sometimes at least – explain the doing' (p. 203). Now, we aren't presently focused on doings that are caused by beliefs, desires, and plans. We are starting 'smaller'. Take a single symbol, perhaps a string of 0's and 1's. One way of viewing what a computer does with the symbol is entirely syntactic – any manipulations are entirely insensitive to any semantic interpretation of the symbol. But in order to escape charges of mimicry, one thing we'd like to be able to correctly say is that the *intelligent* machine doesn't merely push around syntax, but has some kind of sensitivity to the meaning of the symbol and that the meaning is part of what goes on to guide further activity. And this is what someone like Dretske helps provide.[7]

Take some internal state in a system, S; some activity of the system, A; and some feature of the environment the system is in, F. For example, we might have an internal brain state, the raising of an arm, and nutritious fruit hanging from a tree overhead. Or, we might have an internal state in a machine, a downstream state in a machine, and some textual input. Next, suppose that S and F are correlated with each other reliably in such a way that we can correctly say that S *indicates* F – there are lots of ways we might try to spell out the reliable correlation, but for now let's try to avoid getting too deep into those weeds. The important point at present is the stipulation that S is reliably correlated with F. Now, just because S is reliably correlated with F doesn't mean that that fact is relevant to anything that goes on in the brain or machine under consideration. But, let's suppose that the presence of S comes to be *recruited* by the system to serve as a switch or a prompt in the presence of F to cause A. For example, when a creature is in the presence of hanging nutritious fruit, the creature benefits from getting it and eating it and so, over time, the creature is rewarded in situations in which S leads to A *because* S is a reliable indicator of F. According to a theorist such as Dretske (1988), this is exactly what is needed for S to be a *representation* of F. By being an indicator of F that is utilised by the system because it indicates F, S comes to have genuine semantic properties. 'S' means or represents F.

[7] Dretske's work on this front is part of a wider class of work in 'teleosemantics'. I'm focusing on Dretske because the core ideas are particularly clear (in my view any way) and the ideas give one a sense for how teleosemantics works in general as a style of approach. Other seminal work in the area can be found in Millikan (1984) and in Papineau (1987). Recently work in this area that develops these ideas in great detail can be found in Neander (2017) and Shea (2018).

For Dretske, that's how one thing comes to be a representation of another, but what about playing the crucial role of being relevant to the processing? Recall that the worry presented by the octopus was a lack of semantic sensitivity. Dretske has so far provided a story for getting some semantics on the scene, but are they relevant? Does the creature raise its arm *because* of what it represents and does the machine display an output *because* of the meaning of the symbols it manipulates?

The 'because' is slightly complicated. According to Dretske himself, what's uncovered when reflecting on semantics and explanation is a distinctive notion of cause – what he calls a 'structuring cause' – that explains *why* S causes A to occur, but doesn't reveal, as we might put it, *how* things are being pushed around. The idea can be brought out by looking at Dretske's well known example of the thermostat. An engineer designs the thermostat, let's suppose, by placing a bi-metallic strip between two wires. When the temperature changes, the strip coils or uncoils and hence completes the circuit or cuts it. We might ask, 'why did the boiler just turn on?'. One answer is, 'the cooler temperature caused the strip to coil which closed the circuit'. According to Dretske, this is a 'triggering cause' and it is a perfectly good style of explanation. But we might want to know, not by which mechanism did the boiler turn on, but why would a change in temperature have the *causal effect* of *turning on the boiler*? And here, says Dretske, the kind of answer we give is different: the change of shape of the strip was recruited by the designer as a switch to be engaged when the temperature is sufficiently low. Put differently, the engineer designed things in such a way that the bending of the strip represents the temperature and flips the switch *because* of what it represents. According to Dretske, this is an example where one correctly cites semantic content in a structuring-causal explanation.[8]

In the case of the thermostat, we wouldn't be tempted to say that system is exhibiting thinking or cognising, let alone intelligence. And what's said about the thermostat looks like a pretty good thing to say about a classic computer. Let's take a really simple example and at a very low level. Suppose we have a registry with an 8-bit entry and that the first four bits are the operation code and the second four bits pick out a memory location. Let's suppose that we want to describe this entry as saying 'load address 3 into registry 6'. Why might we describe it this way? After all, the bits could just as well have stood for something completely different. The reason it's right to say that the entry means 'load address 3 into registry 6' is because the system was designed in such a way that when the first four bits are processed, a certain procedure kicks off and that procedure is guided by a piece of information that stands for the 6th register. In effect, the system is correctly interpreted as representing things and its activities can be described by citing those representations as causes. The semantics are present (which is a good start), but it seems to be an entirely extrinsic matter. The machine isn't doing anything intelligent, it is merely pushing around symbols that we interpret as meaningful and there are interesting correlations between the semantics and the syntax that open up profitable explanations.

[8] It is worth noting that a deep divide between structuring causes and triggering causes might be an overreaction. Getting into the details would require a long (and inevitably controversial) discussion of causation. But notice that Dretske's story concerning structuring causes features perfectly normally in questions of the form 'What would have happened if…'. As Woodward (2003) puts it, good explanations in science show us why some explanandum phenomenon occurs rather than some alternative outcome, given the conditions being what they are. If we were to see this as a guide to causal explanation more generally, Dretske's structuring causes aren't deeply different from any other causes. See Horgan (1991) and also Rescorla (2012).

But this isn't the only way to get representation and representationally guided activity off the ground. We already saw this above in our fruit gathering creature. In that case, presumably it isn't that some designer put in place some internal structure and process that covaries with the presence of fruit and causes arm raising. Rather, through a process of learning in the individual, and adaptation in the population of which the individual is a member, there are natural pressures that explain the recruitment of a state that covaries with a feature being utilised to generate activity. At this point, we have the beginnings of 'primary' or 'original' intentionality – a naturally occurring semantics. The system, on its own, and not just because of the intentional design of an engineer or the interpretation of someone engaging with the system, is dealing in the semantic. It's because a state or event stands for something that the system continues to utilise the state or event. This is, in my view, and the view of many influenced by Dretske, the very beginnings of mindedness.

LLMs look to fit interestingly into this Dretskean picture. Unlike the predesigned algorithm for moving a value into a register, machine learning opens up the possibility of synthetic, original intentionality. It seems at least in principle possible that a cleverly designed system that can update internal activations and weights might have the right stuff to show sparks of mindedness and perhaps sparks of intelligence. In Dretskean terms, the grounds are in place for semantics that has flexible, causal relevance beyond mere interpretation.

So where have we gotten to? All of that was by way of explaining (i) the importance of semantics or meaning to intelligence, (ii) how semantics could occur in a system in a non-derived-from-the-engineer sort of way and (iii) how the semantic properties of the system can be genuinely relevant when explaining what the system does. These are the features, in my view, that we should be looking for when determining if a machine is intelligent and this is what I think Inner Interpretability can provide for us. And perhaps more exciting still, there are some reasons for thinking that Inner Interpretability has already uncovered some sparks.

## What Can Inner Interpretability Show Us and What Has It Shown Already?

Mechanistic interpretability and Inner Interpretability more generally is a young area of study. Presently, to the best of my knowledge, the most compelling and worked out examples have occurred on truncated versions of GPT-2. For example, 'GPT-2 Small' is a GPT-2 decoder-only transformer-based language model with around 117 million parameters. In comparison, some estimates claim that GPT 4 has about 1.7 trillion parameters. By looking at a truncated model, engineers can work towards deep backward engineering of the algorithms. I can't see an in-principle reason why this methodology should be limited to truncated models, but work must be made manageable. Given my philosophical interests in the present paper, I'm happy with a kind of proof of concept that can be taken forward as the Inner Interpretability field grows and advances.

Let's look at two examples. The first example will help one get a sense of the kind of thing that Inner Interpretability can provide. The second example goes further, suggesting that some AI systems are indeed showing the kernels of cognitive activity and perhaps even the beginnings of intelligence and understanding.

In 'Interpretability in the Wild: A Circuit for Indirect Object Identification in GPT-2 Small', Wang et al (2022) investigate indirect object identification (IOI), aiming to suss out how the model completes the task in terms of internal components. An example of an IOI task is to complete a sentence such as the following: 'After John and Mary went to the shops, John gave a bottle of milk to ___'. The sentence plausibly should end in 'Mary' rather than 'John'.

Wang et al aim to isolate a circuit responsible for completing this task. To do so, they introduce a systematic approach involving information tracing and causal intervention that allows them to isolate a sub-part of the model that causally co-varies with the task in focus. As they describe their methodology:

> In mechanistic interpretability, we want to understand the correspondence between the components of a model and human-understandable concepts. A useful abstraction for this goal is circuits. If we think of a model as a computational graph M where nodes are terms in its forward pass (neurons, attention heads, embeddings, etc.) and edges are the interactions between those terms (residual connections, attention, projections, etc.), a circuit C is a subgraph of M responsible for some behavior (such as completing the IOI task). (p. 3)

Here is an algorithm humans can understand and that would give pretty good results in IOI tasks:

1. Identify all previous names in the sentence (Mary, John, John).

2. Remove all names that are duplicated (in the example above: John).

3. Output the remaining name.

Wang et al argue that they have discovered a circuit that implements this algorithm:

> Our circuit contains three major classes of heads, corresponding to the three steps of the algorithm above:
>
> - Duplicate Token Heads identify tokens that have already appeared in the sentence. They are active at the S2 token, attend primarily to the S1 token, and signal that token duplication has occurred by writing the position of the duplicate token.
>
> - S-Inhibition Heads remove duplicate tokens from Name Mover Heads 'attention. They are active at the END token, attend to the S2 token, and write in the query of the Name Mover Heads, inhibiting their attention to S1 and S2 tokens.
>
> - Name Mover Heads output the remaining name. They are active at END, attend to previous names in the sentence, and copy the names they attend to. Due to the S-Inhibition Heads, they attend to the IO token over the S1 and S2 tokens. (p. 4)

One of the most important contributions of this paper in my mind is the display of a kind of rigorous approach to backward engineering and this is likely to be the main take away for many computer scientists. But I think there is a very interesting philosophical payoff to be taken away as well.

Perhaps GPT-2 Small exhibits original intentionality or even intelligence with respect to identifying *word occurrences*. Put this thought aside and focus on linguistic competence and IOI. I don't think we would be inclined to say that completing the sentence by the above method shows linguistic competence with indirect objects nor with the verb 'to give'. To show that, what we might hope to see is something more like a representation of the syntax of the sentence in such a way that the relations between noun phrases and verb phrases are calculated in rule-governed ways. Or perhaps some kind of retrieval of information concerning the theta-roles of a verb like 'to give', specifically something that gives us reason to think that part of the reason that 'Mary' is the completion rather than 'John' involves something like the information that people don't tend to give things to themselves. Exactly what would satisfy on this front needn't preoccupy us – it depends on an active research agenda in linguistics. The important point for us now is that much like the Chinese Room or the octopus, GPT-2 small is doing something that *looks*, at the level of behaviour, very much like something you or I might do when completing the sample sentence, that is, completing with 'Mary' and not 'John'. But once we get under the hood, we come to think that the method at play is importantly different. Now, being different from the way *we* do it shouldn't be a strike against GPT-2's potential intelligence. As I said earlier, what we might be discovering are sparks of alien intelligence. But when we reflect on something like counting occurrences of words, we quite plausibly think this looks a lot more like mechanical pattern matching than thinking or cognising. And I think this is an exciting result. By looking carefully at the inner workings of the system, we are in a much better position to form hypotheses concerning intelligence than we are when we follow the Black-box methodology. The result here is that we've found a dumb process, but the way we got there is interesting and promising.

It's worth drawing out the methodology in a little more detail given the criticisms of Bubeck et al above. Notice that in my comparison between what GPT-2 small does and what we might think a system exhibiting linguistic competence does, I look at a process and ask whether it *is* an intelligent process or, alternatively, is one that *constitutes* linguistic competence.[9] As was the case earlier, so it is here: I don't want to even attempt to provide necessary and sufficient conditions. But here is a hypothesis: to complete the sentence 'After John and Mary went to the shops, John gave a bottle of milk to ___' with 'Mary' *on the basis of the information* that only John and Mary are salient, John is doing the giving, and people don't usually give things to themselves, *is* a way to complete the sentence with linguistic competence. Maybe there are other ways to do it with linguistic competence, we can leave that open. But, I submit, carrying out 1-3 above is *not* one of those intelligent ways.

Notice how this way of reasoning avoids our earlier worries. First, in Bubeck et al's methodology, we ran the risk of inferring an explanation under the assumption that what goes for humans and animals goes for machines too. I argued that was dubious given important disanalogies. Notice I am not relying on analogical reasoning. Rather, I am making a claim about a process that is or constitutes intelligent activity. Second, there was the worry that inside the black box might be a dumb process rather than an intelligent one. The present methodology is sensitive to this worry by looking inside the box.

Under the present methodology I'm suggesting, we are working with something much closer to the Dretskean idea that intelligent activity is activity driven by and appropriately guided by semantic states. The more specific application is that linguistic competence in IOI tasks is activity governed by the right sorts of representations and their manipulation. GPT-2 Small, for all of its virtues, does not look to be carrying out a process that lives up to the standards of linguistic competence so understood. Of course, there is a great deal of inference to the best explanation afoot (there is nothing wrong with inference to the best explanation!), but notice that we hypothesise about what a certain domain of intelligent activity *consists in* and then look for whether there is good evidence that that's the kind of thing going on. This goes a long way towards closing the explanatory gaps that threaten Bubeck et al's approach.

---

[9] I will put to one side philosophical questions about identity and constitution. At present, what's needed is a tight enough connection to ground the inference from what's observed 'under the hood' to there being an intelligent process on display.

Now, too bad for GPT-2 Small. The IOI task completion doesn't look intelligent in the relevant way. But I don't think for a moment that we should draw a pessimistic conclusion (to the extent that machine intelligence strikes one as something to be optimistic about). I want to turn now to a further study in the Inner Interpretability-style that I think shows some genuine and exciting sparks of intelligence.

In 'Emergent World Representations: Exploring A Sequence Model Trained on a Synthetic Task', Li et al (2023) trained a variant of the GPT model to play legal moves in Othello.[10] The model is a next-move predictor, but, Li et al offers compelling evidence that it spontaneously learns to compute the full board state. Li et al describe this as an 'emergent world representation'.

Abstracting away from various details, the rough idea that Li et al aim to establish is that a modular model of the game board emerges in the system despite the model not being provided with game-board information in training. In comparison to, for example, AlphaGo, no knowledge of board structure or game rules is given to the model. Training to make legal moves is based on lists of moves specifying board positions (such as 'A1' or 'E3'). Othello-GPT is trained to predict the next move given a preceding partial game.

After training, Othello-GPT becomes very good at predicting legal moves. But how is this achieved? From one vantage, the vantage of internal activations, it's complicated token prediction. But Li et al hypothesise that there is a higher level of description in terms of board representations that might be at play.

To investigate this, Li et al proceed in a way not dissimilar to Wang et al, probing[11] for and aiming to isolate internal representations.[12] What Li et al found was evidence of representations with a board-like geometry. That's interesting in and of itself, but of special importance is whether that putative representation is playing a role in the move predictions or is more like the engineer's intentional interpretation discussed earlier.

Li et al extracted the model's activations mid-computation and modified them as if they were modifying a spatial board. Here is a way to think about this: we have a putative representation of a board that itself exploits a board-like geometry. The researchers then change the representation of some targeted board tile into a new state and then plug the modified world representation back in and let it make a prediction with this new world state. If the prediction is the sort of prediction one would expect were one making changes to the represented board, then there is some evidence that such a structure is both present and playing a causal role in the system's prediction. And indeed this is what Li et al found.[13]

[10] See also Li (2023) for summary and discussion.
[11] For more on probing, see Belinkov (2022).
[12] Li et al achieved their results using non-linear probes but could not achieve the results using linear probes. In a recent discussion, Nanda et al (forthcoming) offer reasons for thinking that there is in fact a linear representation of the board state and which can be intervened upon in a linear way. This is important in its own right for determining whether non-linear representations might be atissue. But Nanda et al also make an observation that is exciting for those concerned about reference and indeterminacy: "Our key insight is that rather than encoding the *colours* of the board (BLACK, WHITE, EMPTY), the sequence model encodes the board *relative* to the current player of each timestep (MINE, YOURS, EMPTY)" (p. 2). Let's suppose for a moment Nanda et al have got this right. One is reminded of debates concerning toads and frogs representing flies versus shadows versus nutritious things vs … . Real progress has been made on the best explanation of what the toad represents by devising more and more specific and controlled experiments against the backdrop of more creative hypotheses – see Neander (2017, Ch 5) for a helpful discussion and overview. As more careful and sophisticated interpretability projects are undertaken, we put ourselves in a position to narrow down representational indeterminacy.

I expect that Li et al will garner plenty of discussion and I think some modesty in drawing conclusions is wise. But for my purposes we can take away something valuable. Let's suppose Othello-GPT is doing exactly what Li et al are hoping to establish, namely generating on its own a board-like representation of the Othello board and, further, that the representation lives up to a Dretskean standard by being explanatorily relevant in the downstream predictive processes because of what it represents.[14] Maybe we won't thereby be ready to say that Othello-GPT is an intelligent Othello player, but it is doing something much more exciting than mere lookup. Applying Dretske's model, if Li et al are correct, we have exactly what it takes to establish a real spark of intelligence.

## Guidance for Inner Interpretation

I'd like to briefly point towards a way in which I think philosophy can contribute to Mechanistic Interpretation and Inner Interpretation more generally in light of the foregoing. My sense is that much of what motivates Mechanistic Interpretability researchers is a kind of understanding of the algorithms as at low a level as possible in order to intervene on and manipulate those processes. There is no doubt this is valuable. But the style of work undertaken by Li et al seems to be up to something just a bit different. Li et al are not simply looking for the lowest level algorithms. They are looking for emergent representations with causal/explanatory value that depend on lower level mechanisms. I think this is absolutely crucial for at least three areas of research. First, to make progress on questions about cognitive processes and intelligence, this is precisely the sort of thing we need to be looking for. As I've argued above, one of the important things that differentiates the mimic from the thinker is the manipulation of representations because of what the representations mean. The lowest level of description will not tell us about this representational level and so has no hope of telling us whether intelligence is on display. Second, Li et al's style of approach is crucial for robust interpretability. It's helpful here to compare two descriptions of a bit of human activity. Suppose I see you waving a book around in your office and I find myself confused as to what you are up to. A nearby neuroscience colleague tells me 'Sam's brain is in state B and there's a rush of neurochemical activity of sort N and that's why Sam is waving the book around.' A nearby philosophy colleague tell me, 'There is a fly buzzing around and Sam wants to swat it and believes using a book is a good way to do that.' As with our discussion earlier of Dretske, these are both good explanations, but they are different, non-competing explanations. One tells us about a kind of mechanism at play and the other rationalises Sam's person-level behaviour. Something similar is important when interpreting AI. If (and I really mean IF), we come to think that, say, some LLM is engaged in cognitive activity (and perhaps intelligent cognitive activity to boot), understanding it fully will mean not only explaining the mechanisms but also working towards cognitive/psychological explanations. Those are the sort of facts that will underpin notions such as lying and trust. People often talk about LLMs 'lying' and 'hallucinating', which has clear ramifications for reliability, but we need to know if these are mere metaphors or if there is (or will be) more substance to these indictments. Only a system that engages in the right kind of cognitive activity can lie, assert, and know. Third, there are ethical ramifications. Suppose we are training models to make mortgage decisions and we discover that in two of our models, no Hispanic American applicants have been approved. Already this is a bad result since it diverges from which mortgages would be given by humans following the usual standards. And, crucially, their being hispanic Americans isn't (or shouldn't be) a factor. But just *how* bad is the result? Suppose we did a bit of Inner Interpretability on two models and we find that in one model, the system is showing a sensitivity to a bizarre, financially irrelevant, gerrymandered property that, by some cosmic fluke, correlates perfectly with being an Hispanic American mortgage applicant.[15] The second model is different. It is generating representations of race and ethnicity and using those representations in downstream predictions. Both models are deeply flawed and need to be shut down. But one model is tracking something strange and spurious. The other model is implementing bigotry and racism. This is morally relevant all on its own. Moreover, the bank that hopes to use one of these models will definitely want to avoid using model two.

[13] An important philosophical question arises here that I can only mention in passing, but which constitutes future work: if the system isn't getting any external-world-board inputs, how could it *indicate* a board? A particularly interesting question when thinking about LLMs is how meanings might be anchored. (See Mollo and Milliére (2023) for interesting discussion of what they dub the 'vector grounding problem'). There are a few forthcoming candidate ways grounding might occur: by being parasitic on the input sentences (which are meaningful, though their meanings aren't obviously exploited by the model) and by supervised reinforcement (which, short of installing input sensors, seems like the best bet). But how might reinforcement achieve this end given that the model isn't *deferring* or *intending* to refer the way the supervisor is. An alternative and third way would be to develop an internalist semantics and appeal to unsupervised learning. See Potts (2022) for more on this third way. There is a great deal worthy of investigation here. For present purposes, the argumentative aim is more modest – representations look to arise and be utilised. Precisely what they represent remains up for grabs. In the case of Othello, it might be that what's represented is something like an abstract entity, *the Othello board*.

[14] This is very much in keeping with Potts's (2022) suggestions concerning causal abstraction analysis.

In short, applying intentional/psychological explanations familiar to philosophers of mind and cognitive science to models will be important for interpretability, benchmarking, and ethical alignment. The engineers and the philosophers have a lot to offer each other on these fronts.

[15] See footnote 13.